\documentclass[letterpaper]{article}
\usepackage[preprint]{aaai2027}
\usepackage[hyphens]{url}
\usepackage{graphicx}
\usepackage{caption}
\usepackage{booktabs}
\usepackage{array}
\usepackage{multirow}
\usepackage{amsmath,amssymb}
\usepackage{natbib}
\newcommand{\score}{\textsc{SCORE}}
\newcommand{\rca}{coordinate recovery}
\newcommand{\Rca}{Coordinate recovery}
\newcommand{\topone}{Top-1}
\newcommand{\topfive}{Top-5}
\DeclareMathOperator*{\argmin}{arg\,min}
\DeclareMathOperator*{\argmax}{arg\,max}

\title{\score{}: Subject Coordinate Recovery for\\ Label-Free Cross-Subject EEG-to-Image Retrieval}
\author{Zhenyao Cui, Siyuan Kan, Siyang Li, Ziwei Wang, Dongrui Wu}
\affiliations{School of Artificial Intelligence and Automation, Huazhong University of Science and Technology, Wuhan, China\\
\texttt{zycui@hust.edu.cn}}

\begin{document}
\maketitle

\begin{abstract}
Accurate visual decoding can reveal how the brain represents visual
information and recover perceived content from neural signals such as
electroencephalography (EEG), with potential for neural communication.
However, current EEG-to-image retrieval methods perform far below their
within-subject counterparts for new users without labeled calibration,
limiting real-world deployment. To understand this gap, we analyze EEG
features across subjects and find that different subjects preserve
similar relationships among concepts but express them along different
coordinate directions. We therefore propose Subject Coordinate Recovery
(\score{}), a target label-free framework combining recovery-aware source
training with coordinate alignment at deployment. During training,
\score{} aligns source subject EEG with a common image space and
simulates unseen-subject recovery through source-only episodes. At
deployment, with both encoders frozen, \score{} selects reliable
EEG--image landmarks through hubness-corrected matching and estimates an
orthogonal transformation to recover target EEG coordinates without
source data or target labels. In 200-way retrieval on two public
benchmarks, \score{} outperforms the unadapted baseline for every target
subject and achieves the best overall accuracy. It reaches
53.23\%/83.55\% and 12.01\%/32.16\% \topone{}/\topfive{} on THINGS-EEG2
and Alljoined-1.6M, respectively, surpassing the strongest baselines by
17.45/15.70 and 3.08/4.62 percentage points. Without target labels or
encoder updates, \score{} brings brain-based visual decoding closer to
robust, practical, low-latency deployment across users.
\end{abstract}

\section{Introduction}
\label{sec:intro}

\begin{figure}[!t]
\centering
\includegraphics[width=\columnwidth]{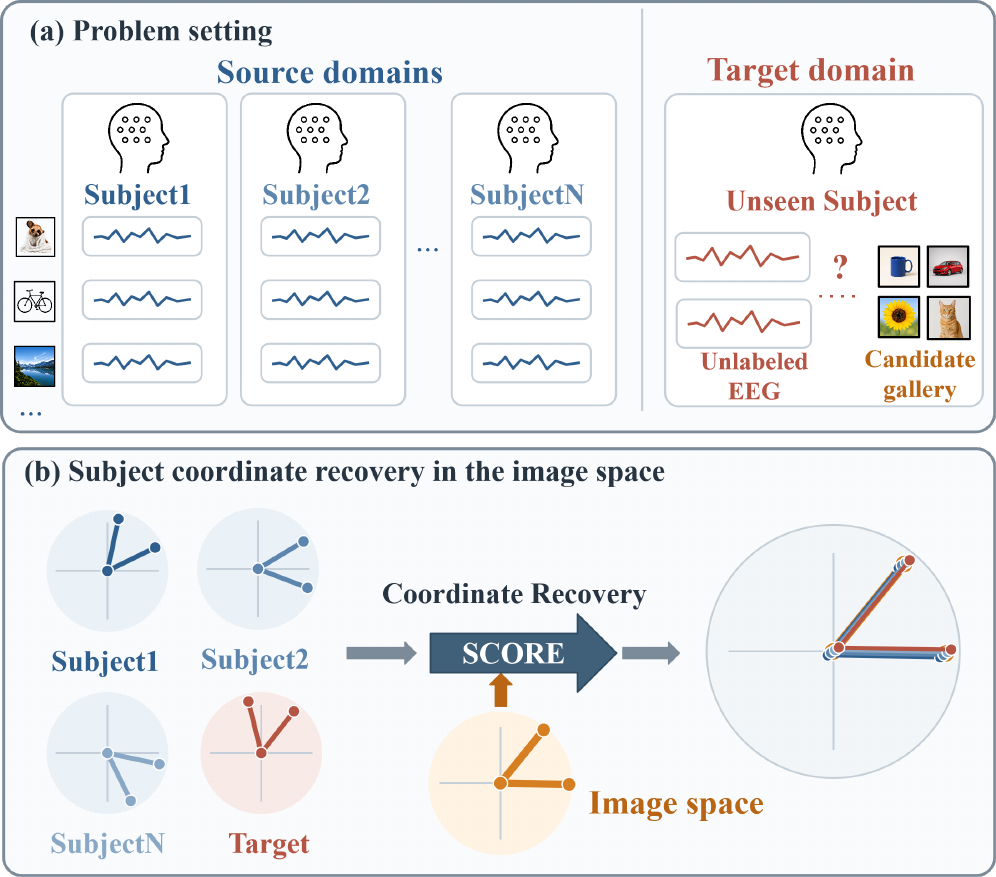}
\caption{Cross-subject EEG-to-image retrieval and the idea behind
\score{}. (a) Source subjects provide labeled EEG--image pairs, whereas an
unseen target subject provides unlabeled EEG and a candidate gallery. (b) Each
subject expresses the same concept relationships along its own coordinate
directions, which \score{} recovers in the image space.}
\label{fig:coordinate-focus}
\end{figure}

Visual decoding aims to recover perceived content from neural
activity~\citep{nice,atms}, offering a way to study visual representations and
support neural communication. Among neural recording modalities,
electroencephalography (EEG) suits practical deployment because
it is non-invasive, portable, and low-cost. EEG visual decoding has been
studied for retrieval and reconstruction~\citep{nice,muse,atms}. In
retrieval, recent studies~\citep{nice,atms,muse,neuroclip}
map EEG and image features into a shared space and align matched pairs
with contrastive learning. On the widely used THINGS-EEG2 benchmark~\citep{thingseeg2},
within-subject approaches have reached 85--91\% \topone{} in zero-shot
200-way retrieval~\citep{hcf,samga}.

However, these results need extensive subject-specific EEG--image
pairs and one model per individual, which is costly and hard to
scale.
Cross-subject retrieval is a more practical alternative. A model trained
on source subjects is applied directly to an unseen subject without
labeled calibration, as shown in Figure~\ref{fig:coordinate-focus}(a).
Existing cross-subject transfer approaches mainly improve generalization by learning
shared source representations or adapting features and scores with
unlabeled target data~\citep{samga,sattc}. Despite these advances,
reported cross-subject \topone{} accuracy on THINGS-EEG2 remains below
35\%~\citep{samga,sattc}. Yet one question remains: why does
EEG-to-image retrieval reach high accuracy within a subject but transfer
so poorly to a new one?

To understand this problem, we analyze the learned EEG representations
across subjects and find that different subjects preserve similar
relationships among image concepts but express them along different
coordinate directions.
Based on this finding, we propose Subject Coordinate Recovery
(\score{}), a framework that combines recovery-aware source
training with label-free coordinate recovery at deployment, as shown in
Figure~\ref{fig:coordinate-focus}(b). During training, \score{} aligns source subject EEG with a
common image space and simulates unseen-subject recovery through
source-only episodes. At deployment, \rca{}
selects EEG--image landmarks under hubness correction and recovers the
target coordinates with a regularized orthogonal map.
\score{} uses no target subject data during training and no target labels
at deployment, making retrieval easier to extend to new users.

On two public benchmarks under 200-way cross-subject retrieval, \score{} reaches $53.23\%/83.55\%$
\topone{}/\topfive{} accuracy on THINGS-EEG2 and $12.01\%/32.16\%$ on
Alljoined-1.6M, surpassing the strongest evaluated baselines by
$17.45/15.70$ and $3.08/4.62$ points, respectively. It
improves both metrics over the unadapted baseline for every held-out
subject on both datasets and generalizes across different EEG
encoders. The recovered map also transfers to an unseen gallery and
remains effective under many distractors, which makes \score{} easy to
scale to new users without target labels or encoder updates.

In summary, our core contributions are as follows:
\begin{itemize}
\item We identify a representational pattern in cross-subject
EEG-to-image retrieval. Different subjects preserve similar concept
relationships but express them along different coordinate directions.

\item We propose \score{}, which brings source subject EEG closer to the
image coordinates during training and recovers a new subject's
coordinates at deployment, using neither source EEG nor target labels
and updating no encoder parameters.

\item \score{} achieves the best overall accuracy among the evaluated
cross-subject algorithms on THINGS-EEG2 and Alljoined-1.6M, improving both
metrics for every held-out subject and generalizing across different
EEG encoders.
\end{itemize}

\section{Related Work}
\label{sec:related}

\paragraph{EEG-to-image retrieval.}
EEG-to-image retrieval projects an EEG trial and candidate images into a
shared space and decodes by similarity matching, which allows test
concepts disjoint from training concepts. NICE established this
formulation on THINGS-EEG2, training an EEG encoder against frozen CLIP
features with a symmetric contrastive loss~\citep{nice}. Subsequent work
improves the source-side representation: ATM extends the embedding to
diffusion reconstruction~\citep{atms}; MUSE preserves multimodal
similarity~\citep{muse}; HCF fuses intermediate visual
features~\citep{hcf}; NeuroCLIP tunes prompts for the visual
encoder~\citep{neuroclip}; and SAMGA builds subject-aware visual
targets~\citep{samga}. These studies reach high accuracy within a
subject, whereas we focus on the cross-subject setting and aim to reduce
the dependence on target labels.

\paragraph{Cross-subject transfer and representation alignment.}
Cross-subject EEG decoding seeks to reduce calibration for a new user~\citep{seizureuda}.
Euclidean Alignment normalizes each subject with a reference covariance
before training~\citep{ea}, while hyperalignment and Riemannian Procrustes analysis align neural
representations across individuals from a common stimulus set or from
class labels~\citep{hyperalignment,rpa}. Source-free subject
adaptation drops the source recordings but still needs target
labels~\citep{sfsaeeg}, and test-time adaptation adapts a frozen model
from unlabeled test data~\citep{tent,ttime}. SATTC targets hubness in
EEG-to-image retrieval and keeps both encoders frozen, calibrating target
features and retrieval scores from unlabeled target EEG alone, combining
subject-adaptive whitening, a score correction based on
cross-domain similarity local scaling, and a structural term over mutual neighbors~\citep{sattc}. These methods bring subjects'
distributions closer together. However, we ask why transfer fails, and use
the answer to recover a new subject's coordinates in the image space,
where retrieval is scored, using neither source recordings nor target
labels.

\paragraph{Unsupervised alignment of two embedding spaces.}
Outside neural decoding, a line of work asks whether two independently
trained embedding spaces can be matched without supervision. For word vectors from two
languages, the correspondence can be recovered by alternating between candidate matches
and an orthogonal transform~\citep{csls,vecmap,wassersteinprocrustes}. We
address a cross-subject problem in neural signals: subjects express the
same concept relationships along different coordinate directions, and an
orthogonal transform recovers much of that difference. The
image space is our anchor: both the source subjects and a new subject are
placed in it, so one fixed target replaces many pairwise alignments.

\section{The \score{} Framework}
\label{sec:method}

\paragraph{Problem setup.}
We consider cross-subject EEG-to-image retrieval. Training uses labeled
EEG--image pairs from source subjects and no target subject data. At
deployment the model receives only unlabeled EEG from a new subject and a
candidate image gallery; no target label is available at any point, which
is what \emph{label-free} means throughout. Formally, the
EEG encoder maps a response $e$ to $q=f_\theta(e)\in\mathbb{R}^d$, and the
image encoder maps a candidate image $v$ to
$g=h_\phi(v)\in\mathbb{R}^d$, where $h_\phi$ projects frozen pretrained
visual features~\citep{samga}. Retrieval returns the gallery
image most similar to the EEG query in this shared space.

\subsection{Cross-Subject Coordinate Diagnosis}
\label{sec:diagnosis}

\begin{figure*}[t]
\centering
\includegraphics[width=\textwidth]{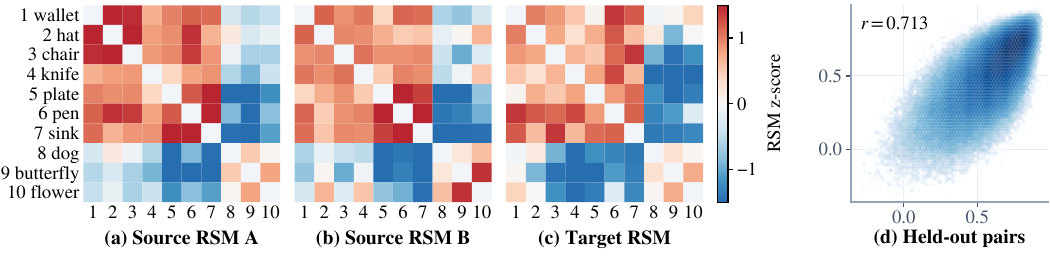}
\caption{Source and target EEG RSMs on encoder-unseen concepts
(THINGS-EEG2), for one LOSO fold. (a)-(c) Two of the nine
source subjects and the held-out target subject, under a shared fixed
z-score scale. (d) All held-out concept pairs for the displayed
source--target pair, with the target subject on the horizontal axis and
source B on the vertical axis ($r=0.713$).}
\label{fig:rsm}
\end{figure*}

\paragraph{Shared concept relationships.}
We first ask whether an unseen target subject preserves the concept
relationships learned from source subjects. For each leave-one-subject-out (LOSO) split of THINGS-EEG2 we train a
separate encoder on one half of the concepts and analyze the held-out
half, so that subject and concepts are both unseen. For each subject,
we construct an EEG representational similarity matrix (RSM) whose entry
$(i,j)$ is the cosine similarity between concepts $i$ and $j$~\citep{rsa}. As shown in Figure~\ref{fig:rsm}, the target and
source RSMs exhibit similar patterns; across target--source pairs the
mean full-RSM correlation is $0.687\pm0.035$. An unseen target therefore
preserves the source concept structure.

\paragraph{Coordinate reorientation.}
Similar concept relationships do not guarantee that coordinates match.
For the held-out concepts, we run three-fold cross-validation. Two folds
fit a ridge map or an orthogonal map from target EEG to source EEG, and
the third evaluates retrieval, so each concept is tested once. The
ridge map can apply any linear transformation, whereas the orthogonal
map only reorients the coordinates and keeps distances and angles
unchanged.

\begin{table}[t]
\centering
\small
\begin{tabular*}{0.82\columnwidth}{@{\extracolsep{\fill}}lcc@{}}
\toprule
Alignment & \topone{} $\uparrow$ & \topfive{} $\uparrow$ \\
\midrule
Direct & $16.89$ & $41.95$ \\
Ridge map & $20.01$ & $47.15$ \\
Orthogonal map & $28.22$ & $59.16$ \\
\bottomrule
\end{tabular*}
\caption{Target-to-source EEG retrieval before and after coordinate mapping on THINGS-EEG2.}
\label{tab:diagnosis}
\end{table}

\begin{figure}[t]
\centering
\includegraphics[width=\columnwidth]{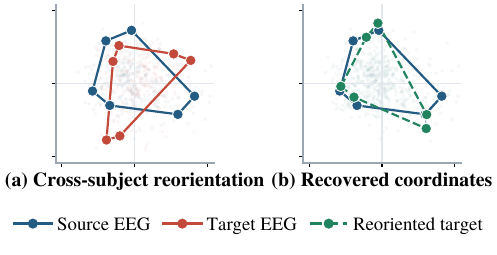}
\caption{Orthogonal coordinate recovery for one target--source pair, projected onto a two-dimensional plane chosen from the fitting concepts.}
\label{fig:reorientation}
\end{figure}

As shown in Table~\ref{tab:diagnosis}, the orthogonal map raises retrieval by $11.33$
\topone{} and $17.21$ \topfive{} points over direct matching, and it also
improves on the ridge map.
Figure~\ref{fig:reorientation} shows the same effect on matched concepts.
Each subject forms a similar shape from the corresponding concepts, yet
the two subjects place that shape under differently oriented coordinate
systems, and the orthogonal map brings the two shapes into close
agreement. These results show that much of the cross-subject gap is a
change of coordinates that preserves distances.

Together, the two diagnostics reveal a specific cross-subject pattern.
First, target EEG preserves source-like concept relationships. Second, the
EEG features of different subjects nevertheless differ substantially, so
direct target-to-source matching is poor, yet an orthogonal map fitted
from known concept pairs largely recovers it. Thus, the information
needed for retrieval survives the change of subject; what blocks it is
the target's coordinate system.

\begin{figure*}[t]
\centering
\includegraphics[width=\textwidth]{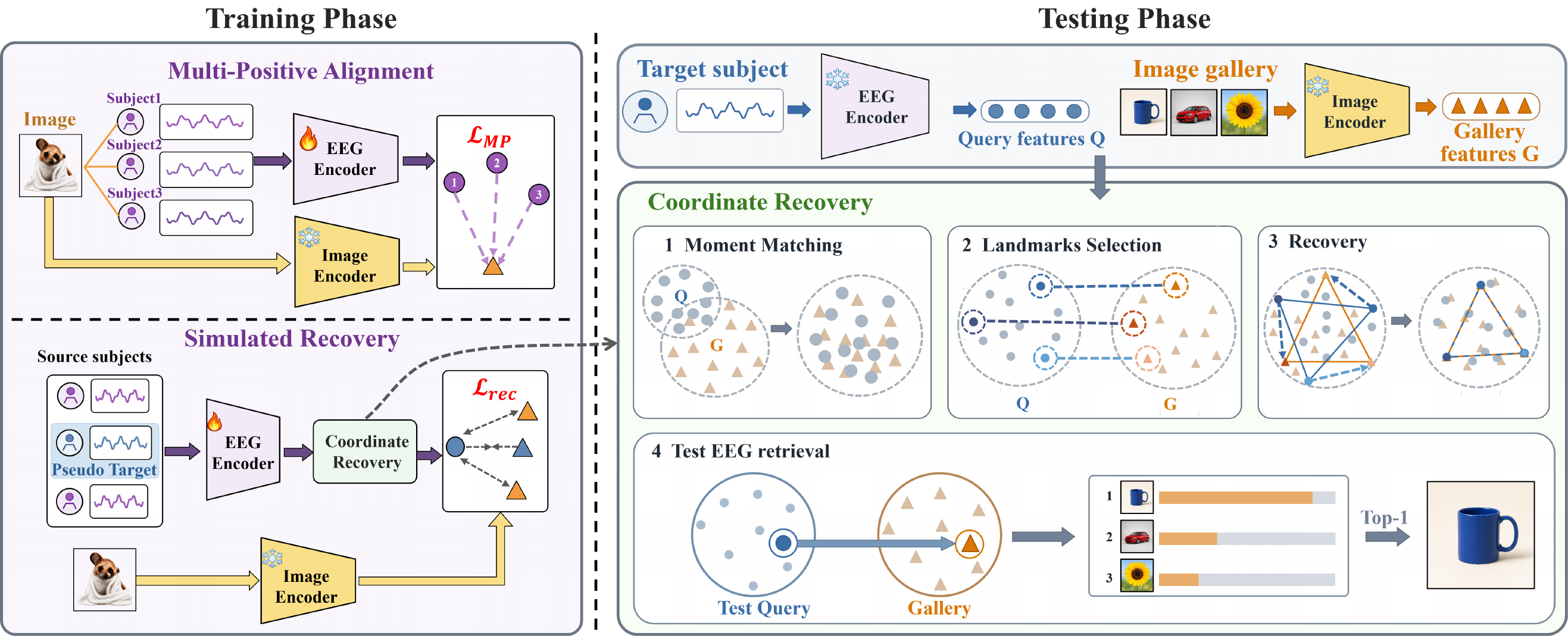}
\caption{Overview of \score{}. Left: recovery-aware source training aligns source subjects and simulates target recovery. Right: at deployment, \rca{} recovers the coordinates of unlabeled target EEG features before image retrieval.}
\label{fig:overview}
\end{figure*}

A direct approach is to align the coordinate systems of the source and
target subjects. However, we align both of them to the
image space instead, for two reasons. First, contrastive source training
has already placed every source subject in the image space, so recovering
a target subject there also aligns it to the source subjects and directly
serves retrieval. Second, it removes the question of
which source subject should serve as the alignment target, and it is
better suited to deployment, where the trained model is available but the
source recordings are not.

This also explains the gap between within-subject and cross-subject
retrieval. Within a subject, the coordinates learned during training
still describe new test concepts, so retrieval stays accurate. A new
subject never took part in that optimization, so it keeps the concept
structure but expresses it in a frame the model cannot read, and accuracy
drops. Serving a new subject therefore comes down to recovering its
coordinates in the image space, replacing a subject-to-subject
difference with an alignment to the gallery already present at
deployment.

\score{} carries this out in two stages: recovery-aware source training
brings source subjects into the image frame using source data only, and
at deployment \rca{} recovers a new subject's coordinates
from
unlabeled target EEG and the candidate gallery, leaving the source
recordings unused (Figure~\ref{fig:overview}).

\subsection{Recovery-Aware Source Training}
\label{sec:training}

\paragraph{Multi-positive alignment.}
Unlike a single-positive contrastive loss, \score{} makes the subject
dimension explicit: EEG responses from different source subjects to the
same image are positives in both retrieval directions, so they are not
pushed apart~\citep{supcon}. The loss therefore pulls the coordinates of
different subjects toward the same fixed image space.

Let $q_i$ and $g_i$ be the $i$-th EEG and image features, and let $y_i$ be
the stimulus index of the $i$-th pair. We define
$\mathcal{P}(i)=\{j: y_j=y_i\}$, $p_{ij}=1/|\mathcal{P}(i)|$ for
$j\in\mathcal{P}(i)$, and $S_{ij}=q_i^\top g_j/\tau$ with temperature
$\tau$. The symmetric multi-positive loss is
\begin{equation}
\label{eq:mp}
\begin{aligned}
\mathcal{L}_{E\rightarrow I}
&= -\frac{1}{B}\sum_i \sum_{j\in\mathcal{P}(i)} p_{ij}
   \log \frac{\exp(S_{ij})}{\sum_l \exp(S_{il})},\\
\mathcal{L}_{I\rightarrow E}
&= -\frac{1}{B}\sum_i \sum_{j\in\mathcal{P}(i)} p_{ij}
   \log \frac{\exp(S_{ji})}{\sum_l \exp(S_{li})},\\
\mathcal{L}_{\mathrm{MP}}
&= \mathcal{L}_{E\rightarrow I} + \mathcal{L}_{I\rightarrow E}.
\end{aligned}
\end{equation}
Here $B$ is the number of EEG--image pairs. The image-to-EEG term supplies
cross-subject supervision because each image is paired with responses from
multiple source subjects.

\paragraph{Simulated coordinate recovery.}
Alignment among source subjects does not ensure the same coordinates for
an unseen subject. We therefore construct source-only recovery
episodes~\citep{episodicdg}. Each mini-batch treats one source subject as a
temporary target, hides its EEG--image matches, and applies the same
recovery steps used at deployment.

After recovery, we reveal the source matches only to compute the loss. Let
$B_t$ be the number of temporary target queries, with recovered query
$\widehat q_i$ and matching image feature $g_{y_i}$. We compute
\begin{equation}
\label{eq:rec}
\mathcal{L}_{\mathrm{rec}}
= -\frac{1}{B_t}\sum_{i=1}^{B_t}
  \log \frac{\exp(\widehat q_i^\top g_{y_i}/\tau_{\mathrm{rec}})}
            {\sum_j \exp(\widehat q_i^\top g_j/\tau_{\mathrm{rec}})},
\end{equation}
where $\tau_{\mathrm{rec}}$ is the temperature. This loss trains the EEG
encoder to preserve the correct retrieval after recovery, while gradients
do not pass through the map.

In the SAMGA implementation used for our main results, multi-positive
alignment replaces the original pairwise contrastive term, and the
recovery loss is added with weight $\lambda_{\mathrm{rec}}$.

\subsection{Label-Free Coordinate Recovery at Deployment}
\label{sec:rca}

At deployment we obtain the target subject's EEG features and the
candidate image features, but no correspondence between them. \Rca{}
estimates the coordinate difference from these unlabeled features in four
steps.

\paragraph{Step 1: Moment matching.}
Before recovering the orientation of the coordinate system, the origins of
the EEG and image coordinate systems must coincide, since an orthogonal
map cannot account for a translation. \Rca{} therefore first matches their
per-dimension statistics, following the practice of re-estimating feature
statistics on the test distribution~\citep{bnadapt}. Let $Q$ and $G$ contain the target EEG and image
features, with $n_q$ and $n_g$ rows in $\mathbb{R}^d$,
\begin{equation}
\label{eq:moment}
\widetilde Q = \frac{Q-\mu_Q}{\sigma_Q+\epsilon}\,(\sigma_G+\epsilon)
             + \mu_G ,
\end{equation}
where $(\mu_Q,\sigma_Q)$ and $(\mu_G,\sigma_G)$ are computed from the EEG
and image features, and $\epsilon$ stabilizes small variances.

\paragraph{Step 2: Landmark selection.}
With the center and scale corrected, the orientation remains, and the
true EEG--image pairs are unknown. We therefore select pseudo-matched
pairs as landmarks, keeping only those informative about the orientation.
However, high-dimensional retrieval suffers from hubness, where a few
images appear among the nearest neighbors of many queries and attract
matches regardless of content~\citep{hubness}, so we score every corrected EEG feature
$\widetilde q_i$ against every image feature $g_j$ with cross-domain
similarity local scaling (CSLS)~\citep{csls},
\begin{equation}
\label{eq:csls}
s^{\mathrm{CSLS}}(\widetilde q_i, g_j)
= 2\cos(\widetilde q_i, g_j) - r_G(\widetilde q_i) - r_Q(g_j),
\end{equation}
where $r_G(\widetilde q_i)$ and $r_Q(g_j)$ are mean similarities to the $k$
nearest features in the other space, penalizing features broadly similar
to many candidates. We retain only mutual nearest-neighbor pairs as
landmarks.

Because EEG has a low signal-to-noise ratio, landmarks should not
contribute equally: a less certain pair should carry less weight. Let
$s_{i,(1)}$ and $s_{i,(2)}$ be the highest and second-highest CSLS scores
for landmark $i$; their margin defines its weight,
\begin{equation}
\label{eq:weight}
\widehat\Delta_i = \max\bigl(s_{i,(1)}-s_{i,(2)},\,\eta\bigr),
\qquad
w_i = \frac{m\,\widehat\Delta_i}{\sum_{j}\widehat\Delta_j},
\end{equation}
where $m$ is the number of landmarks and $\eta>0$ is a small floor.

\paragraph{Step 3: Recovery.}
Step 2 yields $m$ weighted landmarks. Following the cross-subject
coordinate diagnosis, we use them to estimate a $d\times d$
orthogonal map that reorients the target EEG coordinates toward the image
coordinates while preserving distances.

Let $X,Y\in\mathbb{R}^{m\times d}$ contain the landmark rows from
$\widetilde Q$ and their image features. We subtract their
weighted centers $\mu_X$ and $\mu_Y$ to obtain $\widetilde X$ and
$\widetilde Y$, which removes any residual offset before the map is
estimated, and place the weights in a diagonal matrix $W$.

A deployment batch holds a few hundred queries, so at most that many landmarks can be formed, and a
low signal-to-noise ratio leaves fewer still that can be trusted. The
number of landmarks $m$ is therefore much smaller than the feature
dimension $d$, and the alignment term alone is rank-deficient. Its minimizer is then not unique, since $R$ is unconstrained along every
direction no landmark supports. We
therefore add a term that leaves such directions at the identity map $I$.
With nonnegative identity-regularization strength $\lambda$ we
solve
\begin{equation}
\label{eq:rca-procrustes}
R^\star = \argmin_{R^\top R=I}\
\bigl\| W^{1/2}(\widetilde X R - \widetilde Y) \bigr\|_F^2
+ \lambda \bigl\| R - I \bigr\|_F^2 .
\end{equation}
The orthogonal constraint prevents arbitrary scaling. We set
$\lambda=\rho\|\widetilde X^\top W \widetilde Y\|_2$, where $\|\cdot\|_2$
is the spectral norm, so that $\rho$ is dimensionless.

Equation~\eqref{eq:rca-procrustes} has a closed form. Since $R^\top R=I$, the
terms
\[
\operatorname{tr}\bigl(R^\top \widetilde X^\top W \widetilde X R\bigr)
\quad\text{and}\quad
\operatorname{tr}\bigl(R^\top R\bigr)
\]
are constant, and the objective reduces to
\begin{equation}
\label{eq:rca-trace}
\max_{R^\top R=I}\
\operatorname{tr}\bigl(R^\top M\bigr),
\qquad
M = \widetilde X^\top W \widetilde Y + \lambda I .
\end{equation}
Writing the singular value decomposition as $M=U\Sigma V^\top$ gives
$R^\star=UV^\top$~\citep{procrustes}. The identity regularization therefore acts as a
ridge term on the cross-covariance: in directions where
$\widetilde X^\top W \widetilde Y$ carries little energy, $M$ is dominated
by $\lambda I$ and $R^\star$ stays close to the identity.

Because the map is fitted between centered landmark sets, recovery is
completed by subtracting $\mu_X$ from every target feature, applying
$R^\star$, and adding $\mu_Y$,
\begin{equation}
\label{eq:apply}
\widehat Q = (\widetilde Q - \mu_X)\,R^\star + \mu_Y ,
\end{equation}
which places the whole target set, and not only the landmarks, in the
image coordinates.

\paragraph{Step 4: Test EEG retrieval.}
Retrieval proceeds in the recovered space: we recompute the
hubness-corrected similarity of Equation~\eqref{eq:csls} on $\widehat Q$,
so the neighborhood terms come from the recovered features rather than
$\widetilde Q$, and return the highest-scoring candidate for each
query,
\begin{equation}
\label{eq:retrieve}
\hat{\jmath}(i) = \argmax_{j}\ s^{\mathrm{CSLS}}(\widehat q_i, g_j).
\end{equation}
Top-$k$ accuracy counts a query as correct when its paired image is among
the $k$ highest-scoring candidates.

\section{Experiments}
\label{sec:exp}

\subsection{Setup and Comparison Protocol}
\label{sec:setup}

We evaluate \score{} under LOSO protocols on
THINGS-EEG2 and Alljoined-1.6M, which share the THINGS object-concept
stimulus set~\citep{things}, in the standard 200-way
setting~\citep{thingseeg2,alljoined}. Following
prior work~\citep{nice,samga} and the processed release of~\citet{ubp},
on THINGS-EEG2 we use all 63 channels,
segment epochs from 0 to 1000\,ms after onset, baseline-correct
with the 200\,ms pre-stimulus mean, resample to 250\,Hz, apply
multivariate noise normalization (MVNN), and average repetitions of the
same image. Alljoined-1.6M is used as released, under its own preprocessing.

We use SAMGA, a recent strong EEG encoder~\citep{samga}, train each
model for 50 epochs, and report the final epoch. Our strict cross-subject
protocol exposes no target subject data during training, and the
supplement describes SAMGA's reported result and its training protocol.
\Rca{} uses $\rho=0.1$, 12 to 160 landmarks, ten
CSLS neighbors, and a single recovery step.
Recovery-aware training uses
$\lambda_{\mathrm{rec}}=0.03$ and
$\tau=\tau_{\mathrm{rec}}=0.07$.

\begin{table*}[t]
\centering
\small
\begin{tabular*}{\textwidth}{@{\extracolsep{\fill}}llcccc@{}}
\toprule
\multirow{2}{*}{Source training} & \multirow{2}{*}{Deployment method} & \multicolumn{2}{c}{THINGS-EEG2} & \multicolumn{2}{c}{Alljoined-1.6M} \\
\cmidrule(lr){3-4}\cmidrule(lr){5-6}
& & \topone{} $\uparrow$ & \topfive{} $\uparrow$ & \topone{} $\uparrow$ & \topfive{} $\uparrow$ \\
\midrule
\multirow{6}{*}{\textit{Original}}
 & None & $26.22\pm1.08$ & $57.98\pm0.88$ & $6.08\pm0.33$ & $20.55\pm0.79$ \\
 & EA & $27.45\pm0.82$ & $59.37\pm1.07$ & $6.31\pm0.19$ & $20.47\pm0.67$ \\
 & SATTC (SAW) & $30.98\pm0.35$ & $60.85\pm1.00$ & $8.93\pm0.34$ & $27.54\pm1.19$ \\
 & SATTC & $26.12\pm0.45$ & $52.73\pm1.82$ & $6.78\pm0.15$ & $20.70\pm0.98$ \\
 & CSLS & $35.78\pm1.62$ & $67.85\pm1.21$ & $8.04\pm0.35$ & $25.39\pm0.54$ \\
 & \textbf{Coordinate recovery (ours)} & $48.75\pm1.25$ & $80.93\pm0.50$ & $10.67\pm0.25$ & $30.51\pm0.28$ \\
\midrule
\multirow{2}{*}{\textit{Recovery-aware (ours)}}
 & None & $29.33\pm0.57$ & $62.42\pm1.60$ & $9.08\pm0.32$ & $26.78\pm0.14$ \\
 & \textbf{Complete \score{}} & $\mathbf{53.23\pm1.62}$ & $\mathbf{83.55\pm1.13}$ & $\mathbf{12.01\pm0.49}$ & $\mathbf{32.16\pm0.78}$ \\
\bottomrule
\end{tabular*}
\caption{Cross-subject retrieval on two EEG-to-image benchmarks, with the SAMGA encoder throughout. EA is Euclidean Alignment, which normalizes each subject before source training and is then deployed without further adaptation. SATTC (SAW) denotes SATTC's subject-adaptive whitening operator, applied on its own. Values are means and standard deviations across all subjects and three seeds.}
\label{tab:deploy}
\end{table*}

\begin{figure*}[t]
\centering
\includegraphics[width=\textwidth]{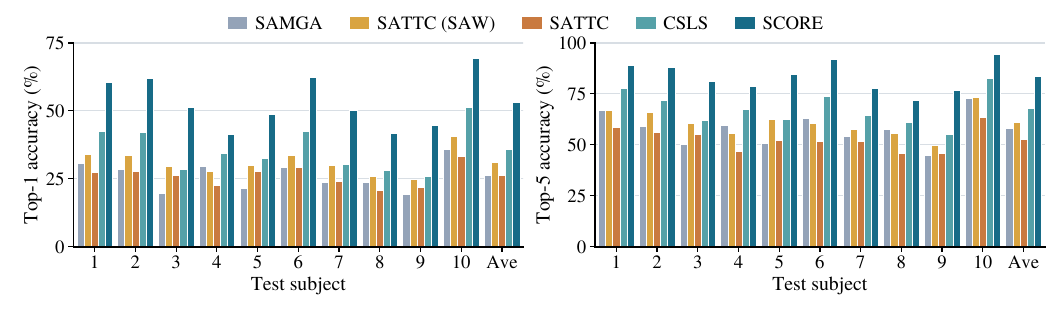}
\caption{Retrieval accuracy for every held-out subject and the mean (THINGS-EEG2), in (a) \topone{} and (b) \topfive{}.}
\label{fig:persubject}
\end{figure*}

\subsection{Main Results}
\label{sec:main}

\begin{figure*}[t]
\centering
\includegraphics[width=0.87\textwidth]{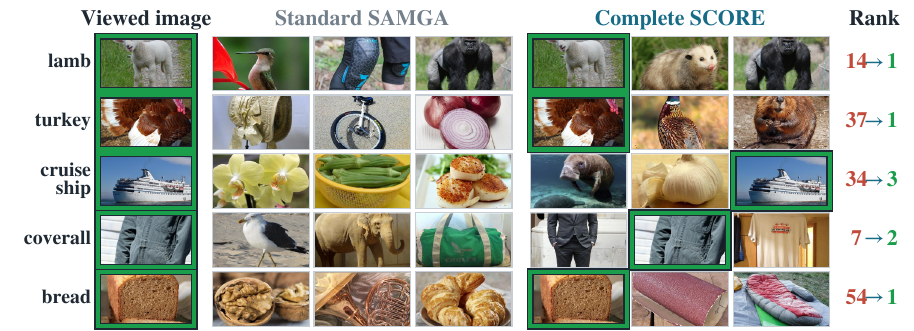}
\caption{Retrieval examples from one held-out subject. Each method shows its top three images, the green border marks the viewed image, and the rightmost column its rank in the 200-image gallery before and after recovery.}
\label{fig:examples}
\end{figure*}

Table~\ref{tab:deploy} reports the main retrieval results. SATTC (SAW)
and CSLS improve retrieval accuracy over the original source-training
baseline. EA adds $+1.23$ and $+0.22$ \topone{}, neither consistent across
folds (sign test, $p=0.585$ and $p=0.568$). \Rca{} yields a clearly larger
improvement on both benchmarks, from the same frozen representation and
without target labels.
The complete SATTC operator adds little to the frozen SAMGA
representation. Unlike SATTC, which was developed on a weaker source
representation, every controlled row here uses a recent strong encoder,
and we report both the complete operator and its strongest
component. Recovery-aware training improves retrieval
further, and complete \score{} achieves the highest accuracy. As shown in
Figure~\ref{fig:persubject}, complete \score{}
improves both \topone{} and \topfive{} accuracy over the unadapted
baseline for every held-out subject on both datasets. In the retrieval
examples of Figure~\ref{fig:examples}, the correct image lies outside the
top three of the unadapted baseline, whereas complete \score{} moves it
into the top three.

\begin{table*}[t]
\centering
\small
\begin{tabular*}{\textwidth}{@{\extracolsep{\fill}}lcccccccc@{}}
\toprule
\multirow{2}{*}{EEG encoder}& \multicolumn{2}{c}{Standard} & \multicolumn{2}{c}{Recovery-Aware} & \multicolumn{2}{c}{Standard + Recovery} & \multicolumn{2}{c}{Complete \score{}} \\
\cmidrule(lr){2-3}\cmidrule(lr){4-5}\cmidrule(lr){6-7}\cmidrule(lr){8-9}
& \topone{} & \topfive{} & \topone{} & \topfive{} & \topone{} & \topfive{} & \topone{} & \topfive{} \\
\midrule
NICE~\citep{nice} & $27.60$ & $60.38$ & $31.08$ & $63.50$ & $50.27$ & $81.05$ & $\mathbf{54.85}$ & $\mathbf{83.75}$ \\
ShallowFBCSPNet~\citep{shallowfbcspnet} & $10.63$ & $32.67$ & $11.37$ & $33.83$ & $28.97$ & $61.45$ & $\mathbf{30.73}$ & $\mathbf{62.92}$ \\
EEGNet~\citep{eegnet} & $15.95$ & $42.80$ & $17.22$ & $45.68$ & $44.28$ & $76.72$ & $\mathbf{46.95}$ & $\mathbf{78.65}$ \\
\bottomrule
\end{tabular*}
\caption{Generalization across EEG encoders on THINGS-EEG2. Values are mean \topone{} and \topfive{} accuracy across three seeds.}
\label{tab:cross-encoder}
\end{table*}

\paragraph{Generalization across EEG encoders.}
Table~\ref{tab:cross-encoder} applies \score{} to three additional EEG
encoders. Adding \rca{} to standard training raises \topone{} by
$18.34$--$28.33$ points, and recovery-aware training gives a further gain
for each encoder. Coordinate recovery therefore attaches to
different EEG encoders as a general component.

\subsection{Analysis of Coordinate Recovery}
\label{sec:analysis}

\begin{table}[t]
\centering
\small
\setlength{\tabcolsep}{3.5pt}
\begin{tabular*}{\columnwidth}{@{\extracolsep{\fill}}llcc@{}}
\toprule
Phase & Configuration & \topone{} $\uparrow$ & \topfive{} $\uparrow$ \\
\midrule
\multirow{3}{*}{\textit{Train}}
 & SAMGA objective & $26.22\pm1.08$ & $57.98\pm0.88$ \\
 & + multi-positive & $28.63\pm1.75$ & $62.02\pm1.73$ \\
 & + simulated recovery & $29.33\pm0.57$ & $62.42\pm1.60$ \\
\midrule
\multirow{4}{*}{\textit{Test}}
 & CSLS ranking & $39.08\pm0.81$ & $71.23\pm0.70$ \\
 & + mean and scale & $43.80\pm1.57$ & $75.68\pm0.58$ \\
 & + recovery & $50.98\pm1.53$ & $80.38\pm0.78$ \\
 & + identity regularization & $\mathbf{53.23\pm1.62}$ & $\mathbf{83.55\pm1.13}$ \\
\bottomrule
\end{tabular*}
\caption{Cumulative ablation of recovery-aware training and coordinate recovery. Each row adds one component to the row above it.}
\label{tab:components}
\end{table}

\paragraph{Training and recovery components.}
Table~\ref{tab:components} separates the contributions of source training
and target recovery. In training, multi-positive alignment gives the
larger gain and simulated recovery a further improvement. At test time,
moment matching improves retrieval, recovery gives the largest additional
gain ($+7.18$ \topone{}), and identity regularization adds $2.25$ points.
Recovering the coordinate directions therefore gives the largest
single test-phase gain, supporting a recoverable coordinate component
across subjects.

\begin{table}[t]
\centering
\small
\begin{tabular*}{\columnwidth}{@{\extracolsep{\fill}}ccccc@{}}
\toprule
Gallery size & Raw & SATTC (SAW) & CSLS & \score{} \\
\midrule
200   & $29.33$ & $35.28$ & $39.08$ & $\mathbf{53.23}$ \\
700   & $15.72$ & $17.98$ & $21.75$ & $\mathbf{29.47}$ \\
1,200 & $12.25$ & $12.80$ & $17.03$ & $\mathbf{22.27}$ \\
1,854 & $9.47$  & $9.92$  & $13.33$ & $\mathbf{17.03}$ \\
\bottomrule
\end{tabular*}
\caption{\topone{} retrieval as training-concept distractors are added to the 200 matching images. Gallery size counts the 200 matching images plus the added distractors. Every column uses the same recovery-aware representation.}
\label{tab:expanded-gallery-main}
\end{table}

\paragraph{Expanded gallery.}
To test whether \rca{} depends on a strict one-to-one query--gallery
correspondence and stays effective once the gallery holds
distractors, we progressively add up to 1,654 training-concept images to
the THINGS-EEG2 gallery, keeping the same target queries and frozen
representations.
Although every method declines as the gallery grows,
Table~\ref{tab:expanded-gallery-main} shows that complete \rca{} remains
best at every size and benefits every held-out subject at the largest
gallery. \Rca{} therefore does not rely on the closed-set assumption, and
\score{} holds its advantage where the gallery contains far more images
than the user has viewed. Methods
that exploit the one-to-one assumption are reported separately in the
supplement.

\begin{table}[t]
\centering
\small
\begin{tabular*}{\columnwidth}{@{\extracolsep{\fill}}llcc@{}}
\toprule
Update & Ranking & \topone{} $\uparrow$ & \topfive{} $\uparrow$ \\
\midrule
None & Cosine & $39.53\pm0.97$ & $75.59\pm0.81$ \\
None & CSLS & $50.41\pm0.60$ & $83.16\pm0.77$ \\
Frozen map & Cosine & $56.35\pm1.14$ & $87.38\pm0.78$ \\
Frozen map & CSLS & $\mathbf{64.16\pm1.59}$ & $\mathbf{91.51\pm0.91}$ \\
\bottomrule
\end{tabular*}
\caption{Transfer of a frozen map between disjoint 100-concept galleries.}
\label{tab:disjoint-gallery-main}
\end{table}

\paragraph{Plug-and-play transfer to an unseen gallery.}
Because the expanded-gallery setting still fits and evaluates \rca{}
within the same query batch, we next split the test concepts into
disjoint Galleries A and B and ask whether a fitted map transfers between
them. \Rca{} estimates the target statistics, landmarks, and map from
Gallery A; the map is then frozen and applied to Gallery B without
refitting.
Table~\ref{tab:disjoint-gallery-main} shows that the frozen map improves
retrieval
for every held-out subject. Since Gallery B shares no EEG queries or
images with Gallery A, the recovered map captures a reusable
subject-to-image relation rather than a fit to specific concepts.

\section{Conclusion}
\label{sec:conclusion}

We investigated why EEG-to-image retrieval transfers poorly across
subjects, and found that different subjects preserve similar concept
relationships while expressing them along different coordinate
directions. Based on this finding, \score{} combines recovery-aware
source training
with label-free recovery of the target coordinates in the image space.
On two datasets and four EEG encoders, it improves retrieval for every
held-out subject, and the recovered map transfers to an unseen gallery,
so a new user can be supported with unlabeled data alone.

However, the method is not without conditions. It still needs landmarks
with a sufficient signal-to-noise ratio, and it still needs a small batch
of unlabeled target samples rather than a single trial. Both conditions
point to the same direction for future work: making the recovered map
reliable from weaker and smaller evidence. We will pursue that direction
next, and we will also extend the framework beyond EEG, to MEG and other
neural recording modalities.

\bibliography{refs}

@inproceedings{nice,
  title={Decoding Natural Images from {EEG} for Object Recognition},
  author={Song, Yonghao and Liu, Bingchuan and Li, Xiang and Shi, Nanlin and Wang, Yijun and Gao, Xiaorong},
  booktitle={International Conference on Learning Representations},
  year={2024}
}

@article{muse,
  title={Mind's Eye: Image Recognition by {EEG} via Multimodal Similarity-Keeping Contrastive Learning},
  author={Chen, Chi-Sheng and Wei, Chun-Shu},
  journal={arXiv preprint arXiv:2406.16910},
  year={2024}
}

@inproceedings{atms,
  title={Visual Decoding and Reconstruction via {EEG} Embeddings with Guided Diffusion},
  author={Li, Dongyang and Wei, Chen and Li, Shiying and Zou, Jiachen and Liu, Quanying},
  booktitle={Advances in Neural Information Processing Systems},
  year={2024}
}

@inproceedings{neuroclip,
  title={{NeuroCLIP}: Brain-Inspired Prompt Tuning for {EEG}-to-Image Multimodal Contrastive Learning},
  author={Wang, Jiyuan and Zhang, Li and Lin, Haipeng and Liu, Qile and Huang, Gan and Li, Ziyu and Liang, Zhen and Wu, Xia},
  booktitle={Proceedings of the AAAI Conference on Artificial Intelligence},
  year={2026}
}

@article{thingseeg2,
  title={A large and rich {EEG} dataset for modeling human visual object recognition},
  author={Gifford, Alessandro T. and Dwivedi, Kshitij and Roig, Gemma and Cichy, Radoslaw M.},
  journal={NeuroImage},
  volume={264},
  pages={119754},
  year={2022}
}

@article{things,
  title={{THINGS}: A database of 1,854 object concepts and more than 26,000 naturalistic object images},
  author={Hebart, Martin N. and Dickter, Adam H. and Kidder, Alexis and Kwok, Wan Y. and Corriveau, Anna and Van Wicklin, Caitlin and Baker, Chris I.},
  journal={PLoS ONE},
  volume={14},
  number={10},
  pages={e0223792},
  year={2019}
}

@inproceedings{bnadapt,
  title={Improving robustness against common corruptions by covariate shift adaptation},
  author={Schneider, Steffen and Rusak, Evgenia and Eck, Luisa and Bringmann, Oliver and Brendel, Wieland and Bethge, Matthias},
  booktitle={Advances in Neural Information Processing Systems},
  year={2020}
}

@inproceedings{csls,
  title={Word Translation Without Parallel Data},
  author={Conneau, Alexis and Lample, Guillaume and Ranzato, Marc'Aurelio and Denoyer, Ludovic and J{\'e}gou, Herv{\'e}},
  booktitle={International Conference on Learning Representations},
  year={2018}
}

@article{hubness,
  title={Hubs in space: Popular nearest neighbors in high-dimensional data},
  author={Radovanovi{\'c}, Milo{\v{s}} and Nanopoulos, Alexandros and Ivanovi{\'c}, Mirjana},
  journal={Journal of Machine Learning Research},
  volume={11},
  pages={2487--2531},
  year={2010}
}

@article{hcf,
  title={Aligning What {EEG} Can See: Structural Representations for Brain-Vision Matching},
  author={Tang, Jingyi and Jiang, Shuai and Su, Fei and Zhao, Zhicheng},
  journal={arXiv preprint arXiv:2603.07077},
  year={2026}
}

@inproceedings{ubp,
  title={Bridging the Vision-Brain Gap with an Uncertainty-Aware Blur Prior},
  author={Wu, Haitao and Li, Qing and Zhang, Changqing and He, Zhen and Ying, Xiaomin},
  booktitle={Proceedings of the IEEE/CVF Conference on Computer Vision and Pattern Recognition},
  year={2025}
}

@article{samga,
  title={Subject-Aware Multi-Granularity Alignment for Zero-Shot {EEG}-to-Image Retrieval},
  author={Jiang, Lin and She, Qingshan and Xu, Jiale and Xu, Haiqi and Wu, Duanpo and Kuang, Zhenzhong},
  journal={arXiv preprint arXiv:2604.17782},
  year={2026}
}

@article{alljoined,
  title={Alljoined-1.6M: A Million-Trial {EEG}-Image Dataset for Evaluating Affordable Brain-Computer Interfaces},
  author={Xu, Jonathan and Nunes, Ugo Bruzadin and Jiang, Wangshu and Ryther, Samuel and Pringle, Jordan and Scotti, Paul S. and Delorme, Arnaud and Kneeland, Reese},
  journal={arXiv preprint arXiv:2508.18571},
  year={2025}
}

@inproceedings{sattc,
  title     = {{SATTC}: Structure-Aware Label-Free Test-Time Calibration for Cross-Subject {EEG}-to-Image Retrieval},
  author    = {Huang, Qunjie and Zhu, Weina},
  booktitle={Proceedings of the IEEE/CVF Conference on Computer Vision and Pattern Recognition},
  year      = {2026}
}

@article{hyperalignment,
  title={A Common, High-Dimensional Model of the Representational Space in Human Ventral Temporal Cortex},
  author={Haxby, James V and Guntupalli, J Swaroop and Connolly, Andrew C and Halchenko, Yaroslav O and Conroy, Bryan R and Gobbini, M Ida and Hanke, Michael and Ramadge, Peter J},
  journal={Neuron},
  volume={72},
  number={2},
  pages={404--416},
  year={2011}
}

@article{rpa,
  title={Riemannian {P}rocrustes Analysis: Transfer Learning for Brain-Computer Interfaces},
  author={Rodrigues, Pedro Luiz Coelho and Jutten, Christian and Congedo, Marco},
  journal={IEEE Transactions on Biomedical Engineering},
  volume={66},
  number={8},
  pages={2390--2401},
  year={2019}
}

@inproceedings{vecmap,
  title={A Robust Self-Learning Method for Fully Unsupervised Cross-Lingual Mappings of Word Embeddings},
  author={Artetxe, Mikel and Labaka, Gorka and Agirre, Eneko},
  booktitle={Proceedings of the 56th Annual Meeting of the Association for Computational Linguistics},
  pages={789--798},
  year={2018}
}

@inproceedings{wassersteinprocrustes,
  title={Unsupervised Alignment of Embeddings with Wasserstein Procrustes},
  author={Grave, Edouard and Joulin, Armand and Berthet, Quentin},
  booktitle={Proceedings of the Twenty-Second International Conference on Artificial Intelligence and Statistics},
  series={Proceedings of Machine Learning Research},
  volume={89},
  pages={1880--1890},
  year={2019}
}

@inproceedings{episodicdg,
  title={Episodic Training for Domain Generalization},
  author={Li, Da and Zhang, Jianshu and Yang, Yongxin and Liu, Cong and Song, Yi-Zhe and Hospedales, Timothy M.},
  booktitle={Proceedings of the IEEE/CVF International Conference on Computer Vision},
  pages={1446--1455},
  year={2019}
}

@article{rsa,
  title={Representational Similarity Analysis---Connecting the Branches of Systems Neuroscience},
  author={Kriegeskorte, Nikolaus and Mur, Marieke and Bandettini, Peter A.},
  journal={Frontiers in Systems Neuroscience},
  volume={2},
  pages={4},
  year={2008}
}

@article{shallowfbcspnet,
  title={Deep Learning with Convolutional Neural Networks for {EEG} Decoding and Visualization},
  author={Schirrmeister, Robin Tibor and Springenberg, Jost Tobias and Fiederer, Lukas Dominique Josef and Glasstetter, Martin and Eggensperger, Katharina and Tangermann, Michael and Hutter, Frank and Burgard, Wolfram and Ball, Tonio},
  journal={Human Brain Mapping},
  volume={38},
  number={11},
  pages={5391--5420},
  year={2017}
}

@article{eegnet,
  title={{EEGNet}: A Compact Convolutional Neural Network for {EEG}-Based Brain--Computer Interfaces},
  author={Lawhern, Vernon J. and Solon, Amelia J. and Waytowich, Nicholas R. and Gordon, Stephen M. and Hung, Chou P. and Lance, Brent J.},
  journal={Journal of Neural Engineering},
  volume={15},
  number={5},
  pages={056013},
  year={2018}
}

@inproceedings{sfsaeeg,
  title={Source-Free Subject Adaptation for {EEG}-Based Visual Recognition},
  author={Lee, Pilhyeon and Jeon, Seogkyu and Hwang, Sunhee and Shin, Minjung and Byun, Hyeran},
  booktitle={2023 11th International Winter Conference on Brain-Computer Interface},
  pages={1--6},
  year={2023},
  doi={10.1109/BCI57258.2023.10078570}
}

@article{ea,
  title={Transfer Learning for Brain--Computer Interfaces: A Euclidean Space Data Alignment Approach},
  author={He, He and Wu, Dongrui},
  journal={IEEE Transactions on Biomedical Engineering},
  volume={67},
  number={2},
  pages={399--410},
  year={2020},
  doi={10.1109/TBME.2019.2913914}
}

@inproceedings{tent,
  title={Tent: Fully Test-Time Adaptation by Entropy Minimization},
  author={Wang, Dequan and Shelhamer, Evan and Liu, Shaoteng and Olshausen, Bruno and Darrell, Trevor},
  booktitle={International Conference on Learning Representations},
  year={2021}
}

@article{procrustes,
  title={A Generalized Solution of the Orthogonal Procrustes Problem},
  author={Sch{\"o}nemann, Peter H.},
  journal={Psychometrika},
  volume={31},
  number={1},
  pages={1--10},
  year={1966}
}

@inproceedings{supcon,
  title={Supervised Contrastive Learning},
  author={Khosla, Prannay and Teterwak, Piotr and Wang, Chen and Sarna, Aaron and Tian, Yonglong and Isola, Phillip and Maschinot, Aaron and Liu, Ce and Krishnan, Dilip},
  booktitle={Advances in Neural Information Processing Systems},
  volume={33},
  pages={18661--18673},
  year={2020}
}

@article{ttime,
  title={T-TIME: Test-Time Information Maximization Ensemble for Plug-and-Play BCIs},
  author={Li, Siyang and Wang, Ziwei and Luo, Hanbin and Ding, Lieyun and Wu, Dongrui},
  journal={IEEE Transactions on Biomedical Engineering},
  volume={71},
  number={2},
  pages={423--432},
  year={2024},
}

@article{seizureuda,
  title={Unsupervised domain adaptation for cross-patient seizure classification},
  author={Wang, Ziwei and Zhang, Wen and Li, Siyang and Chen, Xinru and Wu, Dongrui},
  journal={Journal of Neural Engineering},
  volume={20},
  number={6},
  pages={066002},
  year={2023},
}
\end{document}